# An adaptive Simulated Annealing-based satellite observation scheduling method combined with a dynamic task clustering strategy

Guohua Wu[a*], Huilin Wang[b], Haifeng Li[c], Witold Pedrycz[d,e], Dishan Qiu[a], Manhao Ma[a], Jin Liu[a]

[a] *Science and Technology on Information Systems Engineering Laboratory, National University of Defense Technology, 47 Yanzheng Street, Changsha 410073, Hunan, China.*

[b] *Beijing Institute of Tracking and Telecommunication Technology, Beijing 100094, China*

[c] *School of Civil Engineering and Architecture, Central South University, Changsha 410004, Hunan, P.R. China*

[d] *Department of Electrical & Computer Engineering, University of Alberta, Edmonton, AB T6R 2V4 Canada*

[e] *Warsaw School of Information Technology, Newelska 6, Warsaw, Poland*

**Abstract:** Efficient scheduling is of great significance to rationally make use of scarce satellite resources. Task clustering has been demonstrated to realize an effective strategy to improve the efficiency of satellite scheduling. However, the previous task clustering strategy is static. That is, it is integrated into the scheduling in a two-phase manner rather than in a dynamic fashion, without expressing its full potential in improving the satellite scheduling performance. In this study, we present an adaptive Simulated Annealing based scheduling algorithm aggregated with a dynamic task clustering strategy (or ASA-DTC for short) for satellite observation scheduling problems (SOSPs). First, we develop a formal model for the scheduling of Earth observing satellites. Second, we analyze the related constraints involved in the observation task clustering process. Thirdly, we detail an implementation of the dynamic task clustering strategy and the adaptive Simulated Annealing algorithm. The adaptive Simulated Annealing algorithm is efficient, with the endowment of some sophisticated mechanisms, i.e. adaptive temperature control, tabu-list based revisiting avoidance mechanism, and intelligent combination of neighborhood structures. Finally, we report on experimental simulation studies to demonstrate the competitive performance of ASA-DTC. Moreover, we show that ASA-DTC is especially effective when SOSPs contain a large number of targets or these targets are densely distributed in a certain area.

## 1. Introduction

Since Earth observing satellites (EOSs) exhibit many evident advantages (e.g., a wide field of view without the restriction of territory, etc.) in disaster surveillance, target reconnaissance and intelligence acquiring, they have become indispensable tools in scientific researches, military operations and civil activities. Currently, the accommodation capabilities of existing EOSs are generally still insufficient to meet a large quantity of observation requests coming from various users. Therefore, to rationally make use of scarce satellites resources, efficient satellite observation scheduling methods are necessary. The observation scheduling of satellites amounts to a reasonable arrangement of satellites, sensors, time-windows and sensor slewing angle for observation tasks to maximize the overall observation return, when subject to related constraints. In fact, the satellite observation scheduling problem (SOSP) can be categorized as a kind of multi-dimensional knapsack problem [1], which is NP-hard.

---

*Corresponding author. Tel.:
 E-mail address: guohuawu.nudt@gmail.com

To solve SOSP, exact algorithm, such as the acyclic graph based path search method [2], Depth First Branch and Bound and Russian Doll [1] were firstly developed. In [1], the exact methods were also compared with approximate methods, including Greedy Search or Tabu Search. In general, exact algorithms are only feasible in tackling SOSPs of a relative small size, while approximate algorithms are more suitable to SOSPs of large size. As a result, approximate algorithms have been proposed, such as Lagrangian relaxation and linear search techniques [3], improved greedy algorithm [4], the Iterated Sampling Algorithm and a Squeaky Wheel Optimization [5], Genetic Algorithm [6, 7], Tabu Search [8, 9], Ant Colony Optimization [10-12]. In addition, in [13] a partition-based method was proposed to obtain tight upper bounds of SOSP. It can be found that previous works usually aimed at dealing with specified satellite platforms, for example, Spot series satellites [6, 9, 13], PLEIADES constellation [8], COSMO-SkyMed constellation [14], Landsat 7 satellite [4], ROCSAT-II [3] and FORMOSAT-2 [15].

Task clustering has firstly been demonstrated to be an effectiveness strategy to improve the scheduling efficiency of SOSPs as discussed in our previous work [10]. Task clustering is based on the idea that if multiple observation tasks (corresponding to observation targets) can be executed by the same sensor of the same satellite with the same time-window and sensor slewing angle, these observation tasks can be merged and finished together by a single observation activity. The scheduling process of SOSP with the consideration of task clustering is illustrated in Fig.1. Some potential advantages of task clustering can be summarized as (1) it enables a satellite to finish more tasks at the cost of fewer sensor opening times. (2) It may enable some previously mutual exclusive tasks to be finished simultaneously. (3) It may enable a satellite to accomplish more tasks while reducing sensor slewing time and range, thus resulting in less energy consumption.

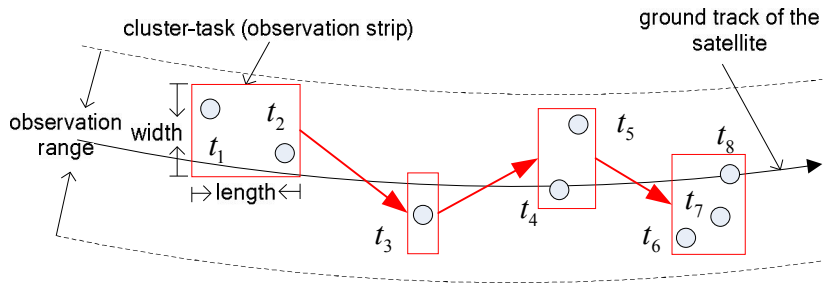

Fig.1. Illustration of the scheduling process of SOSP considering observation task clustering.

Considering the high complexity of both task clustering and satellite scheduling, we began with the presentation of a two-phase based scheduling framework, which was experimentally demonstrated to be superior to some other state-of-the-art satellite observation scheduling algorithms [10]. This framework separates the task clustering operation from the task scheduling process. Obviously, the task clustering strategy in the two-phased based scheduling framework is static, that is, once the cluster-tasks have been generated in the first task clustering phase, they will not be changed in the latter second task scheduling phase. Since cluster-tasks are obtained without careful consideration of the following task scheduling process, the potential of the task clustering have not been fulfilled to the highest extent.

To overcome the shortcoming of the static task clustering strategy, in this work, we present an adaptive Simulated Annealing algorithm integrated with a dynamic task clustering strategy (or ASA-DTC, for short) for solving SOSPs. The corresponding task clustering operations are embedded into the neighborhood search process of the ASA-DTC.

The proposed novel adaptive Simulated Annealing is augmented by being combined with three sophisticated search mechanisms. The first mechanism is an adaptive temperature control function instead of a constantly descending function used to realize the annealing schedule. The second one is the use of a tabu-list to avoid short-term revisiting of the same solution component. The third one is the intelligent combination of two neighborhood structures. In the solution searching process, the neighborhood structures are selected automatically according to their past performance. It is worth stressing that the dynamic task clustering strategy is essentially integrated into the neighborhood structures. The employed mechanisms can effectively improve the performance of the "conventional" Simulated Annealing, which otherwise may easily suffer from premature and be time-consuming in finding high-quality solutions.

In addition, we introduce a new task clustering constraint, i.e. the resource consumption constraint. This constraint requires that if multiple tasks can be finished independently with less resource consumption, then these tasks should not be clustered. In this study, resources are referred to as energy and memory storage. The resource consumption constraint helps avoid useless cluster-tasks, providing more exact guidance for the task clustering operation.

The major contributions of this paper are summarized as follows.

(1) We propose an adaptive Simulated Annealing based scheduling algorithm integrated with the dynamic task clustering strategy (ASA-DTC) for EOSs. The integration is realized by performing task clustering dynamically.

(2) We improve the dynamic task clustering by introducing a new resource consumption constraint, which can prevent from generating useless cluster-tasks to save resources.

(3) We employ three novel mechanisms incorporated into ASA-DTC, i.e., adaptive temperature control, tabu-list based revisiting avoidance technique, and the intelligent combination of different neighborhood structures, to improve the existing Simulated Annealing algorithm. Especially, the method for intelligent combination of different neighborhood structures is first presented in this study.

(4) We conduct extensive experimental simulations and comparative analysis to demonstrate the efficiency of ASA-DTC.

Point targets are considered in this paper. One point target is related to an observation task. We pay attention on the issue of dynamic task clustering and its integration with the adaptive Simulated Annealing scheduling algorithm. Sensors of satellites considered in our study are able to slew laterally. Primary constraints are taken into account, including maximum sensor opening (slewing) times, memory storage capacity, onboard energy capacity and setup time between consecutive tasks.

The paper is structured as follows: Section 2 reviews related work in literatures. In section 3 we present an integer programming model for SOSP. Section 4 analyzes related constraints involved in task clustering. Section 5 introduces the adaptive Simulated Annealing based scheduling algorithm, which is combined with a dynamic task clustering strategy. Section 6 reports on experimental results and offers a detailed performance analysis. Section 7 concludes this paper identifying future research directions.

## 3 A mathematical model of SOSP

The SOSP can be formally described in a form of an integer programming model. As shown in Fig. 1, the satellite observation scheduling aims to arrange an observation sequence for each orbit of satellites to maximize the observation return, subject to corresponding constraints. Different orbits of satellites can be viewed as the same kind of resource equipped with certain observation capabilities. For simplicity, we organize all the orbits together to translate a SOSP with multi-satellite and multi-orbit into a SOSP with multi-orbit. Let $O = \{o_j \mid j = 1,2,\ldots M\}$ be the collection of orbits, where $M$ denotes the total number of orbits of all satellites.

Let $T = \{t_i \mid i = 1,2,\ldots N\}$ denote a collection of the observation tasks. Decision variable $x_{ij}$ indicates whether task $t_i$ is scheduled to be executed on orbit $o_j$. We set

$$x_{ij} = \begin{cases} 1, & \text{if task } t_i \text{ is scheduled to be executed in orbit } o_j, \\ 0, & \text{otherwise.} \end{cases}$$

If orbit $o_j$ have no visibility for task $t_i$, $x_{ij}$ is set to zero.

Some relevant notations used in the model are defined as follows.

| | |
|---|---|
| $TW_i = [ts_i, te_i]$ | time-window of task $t_i$ |
| $p_i$ | weight associated with task $t_i$. |
| $y_{ih}$ | 1-0 variable indicating whether task $t_h$ will be executed after $t_i$ |
| $W_j$ | memory storage capacity at orbit $o_j$ |
| $w_j$ | memory consumption rate by an observation at orbit $o_j$ |
| $E_j$ | energy capacity at orbit $o_j$ |
| $eo_j$ | energy consumption rate by an observation at orbit $o_j$ |
| $es_j$ | energy consumption rate by a sensor slewing operation at orbit $o_j$ |
| $v_j$ | sensor slewing velocity at orbit $o_j$ |
| $a_j$ | required setup time for opening and calibrating the sensor at orbit $o_j$ |
| $\theta_i$ | slewing angle to observe task $t_i$ at orbit $o_j$ |

$c_j$                      maximum times for a satellite opening its sensor at orbit $o_j$

The objective of the scheduling is to maximize the profit expressed by the sum of weights of scheduled tasks:

$$\max \sum_{j=1}^{M} \sum_{i=1}^{N} x_{ij} \cdot p_i . \tag{1}$$

Complex constraints need to be taken into account in the scheduling process. The constraints considered in this study are described as follows [10].

$$\sum_{j=1}^{M} x_{ij} \leq 1, \quad i = 1, 2, \ldots, N \tag{2}$$

$$ts_h - te_i \geq a_j + (|\theta_i| + |\theta_h|)/v_j, \quad i, h = 1, 2, \ldots, N, \quad y_{ih} = 1, \quad j = 1, 2, \ldots, M, \tag{3}$$

$$\sum_{i=1}^{N} x_{ij} \cdot eo_j \cdot (te_i - ts_i) + \sum_{i=1}^{N} \sum_{h=1}^{N} x_{ij} \cdot x_{hj} \cdot y_{ih} \cdot es_j \cdot ((|\theta_i| + |\theta_h|)/v_j) \leq E_j, \quad j = 1, 2, \ldots, M \tag{4}$$

$$\sum_{i=1}^{N} x_{ij} \cdot w_j \cdot (te_i - ts_i) \leq W_j, \quad j = 1, 2, \ldots, M \tag{5}$$

$$\sum_{i=1}^{N} x_{ij} \leq c_j, \quad j = 1, 2, \ldots, M \tag{6}$$

Constraint (2) indicates that each task requires only one observation. Constraint (3) shows that enough setup time is required for a satellite to open its sensor and calibrate the pointing angle of its sensor. Constraint (4) means that at each orbit $o_j$, the energy to be consumed should not exceed the maximum capacity. The energy is mainly consumed by observation and sensor slewing activities. Constraint (5) describe that at each orbit $o_j$, the memory storage to be consumed should not exceed the maximum capacity. Constraint (6) means that a satellite needs to open its sensor before performing each observation task (i.e., the sensor opening time corresponds to an observation task). At each orbit $o_j$, the number of sensor opening times (denoted by $c_j$) for a satellite is limited due to physical constraints and the limitation of onboard energy. In other words, a number of observation tasks can be executed at each orbit. As a result, the number of execution tasks should not be larger than the number of sensor opening times.

It should be noted that tasks in this model can be the initial tasks or cluster-tasks. The cluster-task and the task clustering strategy will be introduced in the following section.

**4. Task clustering constraints**

The prerequisite for combining multiple tasks into a cluster-task is that these tasks can be finished with the same slewing angle and time-window, which constraints the task clustering process. Furthermore, a rational task clustering operation is also determined by the resource consumption constraint which is used to evaluate whether it is worthy to conduct the task clustering.

4.1 Slewing angle and time-window calculation for a cluster-task

A target is related to a task here. Since the tasks clustering constraints are related to the time-window and sensor slewing angle, we complete calculations for the time-window and sensor slewing angle of a cluster-task before introducing the task clustering constraints. As a sensor exhibit a field of the view, a satellite can observe a target within a certain range of sensor slewing angle. For example, the sensor slewing angle range for observing target $t_i$ is $[\underline{\theta_i}, \overline{\theta_i}]$. One of the prerequisites so that multiple tasks can be clustered together is that they can be completed by a sensor with the same slewing angle, that is, the corresponding multiple targets can be covered by a common observation strip. Therefore, if multiple tasks, say $t_1, t_2, \ldots, t_n$, can be clustered into cluster-task $t_u$, they should satisfy the requirement,

$$[\underline{\theta_1}, \overline{\theta_1}] \cap [\underline{\theta_2}, \overline{\theta_2}] \cap \cdots \cap [\underline{\theta_n}, \overline{\theta_n}] \neq \varnothing. \tag{7}$$

Let $\theta_u$ denote the sensor slewing angle for executing cluster-task $t_u$ and $[\underline{\theta_u}, \overline{\theta_u}]$ stands for the possible sensor slewing angle range for executing cluster-task $t_u$. Then we have

$$[\underline{\theta_u}, \overline{\theta_u}] = [\underline{\theta_1}, \overline{\theta_1}] \cap [\underline{\theta_2}, \overline{\theta_2}] \cap \cdots \cap [\underline{\theta_n}, \overline{\theta_n}]. \tag{8}$$

Here we let

$$\theta_u = \frac{1}{2}(\underline{\theta_u} + \overline{\theta_u}). \tag{9}$$

We also need to calculate the time-window for each cluster-task, permitting the satellite to finish the component tasks of this cluster-task in a common temporal interval. The time-window of $t_u$ is denoted by $TW_u = [ts_u, te_u]$. To calculate $TW_u$ for $t_u$, we let

$$ts_u = \min\{ts_l \mid l = 1, 2, \ldots, n\}, \quad te_u = \max\{te_l \mid l = 1, 2, \ldots, n\}. \tag{10}$$

Note that the weight of a cluster-task is just the sum of weights of its component initial tasks.

4.2 Task clustering constraint

Since multiple tasks in a cluster-task have to be finished with the same sensor slewing angle, formula (7) describes the slewing angle related constraint for task clustering. In addition, The longest duration (denoted by $\Delta T$) allowed for a continuous observation is limited due to the characteristic of the sensor in use. Therefore, the time-window $TW_u$ of cluster-task $t_u$ should satisfy

$$te_u - ts_u \leq \Delta T. \tag{11}$$

Task clustering can reduce the sensor slewing angle, enabling satellites to save energy. However, it is not advisable to cluster two observation tasks if there is a large temporal gap between the time-windows of these two tasks. This is because if such two tasks were clustered together, the satellite would perform redundant observation in the time gap. For example, the observation in temporal gap $[te_1, ts_2]$ is redundant. Large $[te_1, ts_2]$ will lead to more consumption of energy and memory storage. Therefore, we should be more prudent when applying the task clustering strategy.

Noticeably, if a satellite can finish previously exclusive tasks (because of constraint (3)) after task clustering, it is worth clustering these tasks. However, if a satellite can finish multiple tasks one after another and consume fewer resources in this manner, the clustering operation on these tasks will not be recommended even though these tasks in fact could be clustered from the view of time-window (constraint (7)) and sensor slewing angle

(constraint (11)).

If task $t_i$ and $t_h$ can be possibly finished at orbit $o_j$ successively and separately. And they also satisfy constraint (7) and (11) to be clustered into $t_u$. Then to finish task $t_i$ and $t_h$ without task clustering, the consumed energy is

$$en_{ih} = eo_j \cdot ((te_i - ts_i) + (te_h - ts_h)) + es_j \cdot ((|\theta_i| + |\theta_h|)/v_j). \tag{12}$$

The consumed memory is

$$wn_{ih} = w_j \cdot ((te_i - ts_i) + (te_h - ts_h)). \tag{13}$$

In contrast, to finish task $t_i$ and $t_h$ with task clustering, the consumed energy is

$$ec_{ih} = eo_j \cdot (te_u - ts_u) + es_j \cdot (|\theta_u|/v_j). \tag{14}$$

While the consumed memory in this case is

$$wc_{ih} = w_j \cdot (te_u - ts_u), \tag{15}$$

where $te_u - ts_u = te_h - ts_i$ and $\theta_u = (\theta_i + \theta_h)/2$.

Considering that task $t_i$ and $t_h$ can be finished separately or together as a cluster-task, according to (12)-(15), if there exist

$$\alpha_j \cdot ec_{ih} + \beta_j \cdot wc_{ih} < \alpha_j \cdot en_{ih} + \beta_j \cdot wn_{ih}, \tag{16}$$

we say the task clustering can save resources and is effective. There $\alpha_j$ and $\beta_j$ stand for the weight of energy and memory storage, respectively. We let $\alpha_j$ equal to the ratio between the consumed energy (denoted by $ConE_j$) and maximum energy capacity at orbit $o_j$, namely, $\alpha_j = ConE_j/E_j$. It is clear that $\alpha_j$ will gradually increase with the energy being consumed. Similarly, $\beta_j$ stands for the ratio between the consumed memory storage (denoted by $ConW_j$) and maximum memory storage, i.e. $\beta_j = ConW_j/W_j$.

Noticeably, formula (16) reflects a consumption constraint. If inequality (16) is not satisfied, it would be better to finish task $t_i$ and $t_h$ separately instead of finishing them as a cluster-task, because to finish task $t_i$ and $t_h$ separately will require resources (evaluated by $\alpha_j \cdot en_{ik} + \beta_j \cdot wn_{ik}$). Therefore, we can say that (7) and (11) guarantee the feasibility of task clustering while (16) guarantees the effectiveness of task clustering.

We have detailed the constraints of task clustering. Once a number of tasks have been clustered, these tasks will be treated and finished as on observation task. How and when conduct the task clustering operation in the scheduling process will be given in the following section.

**5 Adaptive Simulated Annealing**

Simulated Annealing (SA) is based on an analogy of thermodynamics with the way metals cool and anneal. SA has received significant attention in the last two or three decades to solve optimization problems, where a global minimum/maximum is hidden among many local minima/maxima [16]. SA has been widely applied to scheduling problems, such as job shop scheduling [17], ship routing [18], course timetabling [19], knapsack packing problem [20], single-machine scheduling [21] and flow shop scheduling [22]. We choose Simulated Annealing as a search method in this study not only because of its remarkable efficiency shown in solving scheduling problems, but also it is easy to integrate the dynamic task clustering strategy.

In addition, although SA is a robust technique for solving hard combinatorial problems and can converge to the optimal solution given sufficient computation time, it also suffers from premature convergence while being time-consuming in finding high-quality solutions. In this study, we incorporate into SA three adaptive mechanisms, including adaptive temperature control, tabu-list based revisiting avoidance mechanism, and the intelligent combination of two neighborhood structures. We will be referring to the proposed new SA as an adaptive Simulated Annealing (ASA).

To fully describe problem-specified ASA, we define some notations.

$S$ is a schedule,

$S'$ is a neighboring schedule of $S$.

$F$ is the set of overall scheduled tasks.

$F_j$ is the set of scheduled tasks to be executed at orbit $o_j$. The scheduled tasks in $F_j$ can be initial tasks or cluster-tasks. Initial tasks included in a cluster-task are referred to as the component tasks of this cluster-task.

$U$ denotes the collection of the overall unscheduled initial tasks.

$U_j$ denotes the collection of the unscheduled initial tasks owning time-window at orbit $o_j$.

Obviously, there exist $U = \bigcup_{j=1}^{M} U_j$, $F = \bigcup_{j=1}^{M} F_j$ and $T = F \cup U$. Tasks in $U_j$ and $U$ are ranked in descendant order according to their weights. In addition, for an unscheduled task $t_k$, it may have observation opportunities in different orbits.

$Opt_k$ represents the collection of observation opportunities of task $t_k$,

$Opt_{k,j}$ is an element in $Opt_k$, described as $\{TW_{k,j}, \theta_{k,j}, o_j\}$, which means that $t_k$ can be finished in $o_j$ with time-window $TW_{k,j}$ and slewing angle $\theta_{k,j}$.

5.1 Adaptive neighborhood structure

Diversity of solution is an important factor to improve the performance of an intelligent stochastic algorithm. Various neighborhood structures can help generate new solutions in different ways, such that they are able to produce more diversified solutions, attempting to produce an optimal solution. In this study, we developed two kinds of neighborhood structures, namely, the insertion and removal neighborhood, and the task immigration neighborhood. The cluster tasks are continuously changing as a result of the moving around component tasks due the conduction of the neighborhood operation. Note that schedule does not allow for constraint violations, so the conflicting tasks are not scheduled in the both neighborhood structures. Moreover, an adaptive mechanism to select an appropriate neighborhood structure dynamically during the run of ASA-DTC is first propose in this work.

(1) Insertion and removal neighborhood

The insertion and removal neighborhood (also known as a swap neighborhood) is a well-known strategy to generate new neighboring solutions. This neighborhood structure attempts to select an unscheduled task and insert it into the current schedule by clustering with a scheduled task or being an independent scheduled task. Then tasks conflicting with the new inserted task will be removed from the schedule (by checking constraint (3)-(6)). The procedure of this neighborhood structure is outlined as follows.

**Step 1:** Select an unscheduled task $t_k$ with the highest weight from set $U$. Besides, $t_k$ is not banned by the tabu-list to be discussed latter. Go to Step 2.

**Step 2:** Test whether there exists any observation opportunity $Opt_{k,j}$ in $Opt_k$ to allow task $t_k$ to be clustered with a scheduled task in a $F_j$ according to constraint (7), (11) and (16). If it does, go to Step 3; otherwise go to Step 6. Such kind of observation opportunity is referred to as clustering-insertion observation opportunity. The other observation opportunities are called isolated-insertion observation opportunities.

**Step 3:** Assess if there is only one clustering-insertion observation opportunity. If yes, go to step 4; otherwise go to step 5.

**Step 4:** Insert $t_k$ into the only clustering-insertion observation opportunity and cluster it with a scheduled task in $F_j$. Go to Step 7.

**Step 5:** Among multiple clustering-insertion observation opportunities, choose the observation opportunity that is at an orbit $o_j$ owning more resources, evaluated by $(E_j - ConE_j)/E_j + (W_j - ConW_j)/W_j$ (refer to formula (16)). Insert $t_k$ into the selected observation opportunity by clustering it with any scheduled task. Go to Step 7.

**Step 6:** Choose an isolated-insertion observation opportunity that is at an orbit $o_j$ owning more resources, and insert $t_k$ into the selected observation opportunity in an independent manner (without clustering with other scheduled tasks). Go Step 7.

**Step 7:** Assess whether the insertion meets related constraint (3), (4), (5) and (6). Let $o_j$ denote the orbit into which $t_k$ is inserted. If constraint (3) is not satisfied, remove conflicting initial tasks or related component tasks of conflicting cluster-tasks. Afterwards, if constraint (4), (5) and (6) are still not met, remove previously scheduled tasks one by one until all these constraints are satisfied. The scheduled tasks with higher resource consumption and lower weight are preferred to be removed firstly. Note that for the sake of (16), only the component tasks positioned in the end points (with earliest or latest time-window) of cluster-tasks could be removed first each time.

In the process above, we classified the insertion with clustering operation as clustering insertion. And the insertion without clustering operation is categorized as isolated insertion.

(2) Task migration neighborhood

Some tasks can possibly be executed at multiple orbits, i.e., these tasks possess more than one time-window (one time-window is related to a single observation opportunity) in these orbits simultaneously. To enhance the diversity of solutions, we consider immigrating scheduled initial tasks among their different observation opportunities. This neighborhood structure attempts to transfer a scheduled task from its current position (also related to observation opportunity) to another position of the schedule. In the task migration process, possible unscheduled task could be inserted into the schedule because of the freed resources accompanied in the migration. The conflicting scheduled tasks will be removed from the schedule. In this process, the procedure of this neighborhood structure is outlined as follows.

**Step 1:** The scheduled initial task $t_k$ in $F$ possessing more setup time conflicting tasks are selected first. Here, setup time conflicting tasks of $t_k$ are resulted from constraint (3).

**Step 2:** We first remove task $t_k$ from its current scheduled position. Then similar to the insertion and removal neighborhood structure, we insert $t_k$ into another observation opportunity by clustering insertion or isolated insertion operation.

**Step 3:** As the removal of $t_k$ from its previous position can free of some resources, thereby we try to insert the unscheduled setup time conflicting tasks of $t_k$ one by one until the constraint (3) has been violated. As usual, in this process, the setup time conflicting task with higher weight will be selected as the first one.

**Step 4:** Like in Step 7 involved in the insertion and removal neighborhood structure, related conflicting tasks (due to constraint (3)-(6) here) generated in the task migration will be removed to guarantee the feasibility of the obtained solution.

From the two neighborhood structures discussed above, we can see that the task clustering strategy is integrated into the insertion operation and dynamically executed in the neighborhood search process of ASA-DTC. It should be noted that once the cluster-task has been changed, including the removal or addition of component tasks, its slewing angle, time-window and weight should be recalculated according to (9) and (10).

(3) Adaptive combination of the proposed two neighborhood structures

Instead of assigning each neighborhood structure a predefined constant execution probability, we propose a new adaptive combination mechanism which selects the two neighborhood structures dynamically in the solution searching process of ASA-DTC in terms of their past performance. The adaptive mechanism is based on the assumption that the most successful neighborhood structure used in the recent iterations may also be successful in future several iterations.

Let $pro_1$ and $pro_2$ stand for the execution probabilities determining the selection of the first or the second neighborhood structure to generate new neighboring solutions. Initially, we assign equal probabilities to both structures, namely $pro_i = 0.5$, for $i = 1, 2$. Then after a certain number of iterations $Itr$, the following rule is used to update the execution probability of the $i$ th neighborhood structure,

$$pro'_i = \eta \cdot pro_i + (1-\eta) \cdot \frac{Suc_i}{Sel_i}, \tag{17}$$

$$pro_i = pro'_i \Big/ \sum_{i=1,2} pro'_i, \tag{18}$$

where, $\eta$ is an inertial weight to track and keep the previous selection probability of the neighborhood structure. $(1-\eta)$ indicates the weight of learning from the latest past performance of the neighborhood structure when update its current selection probability. $Sel_i$ stands for the times for selecting the $i$ th neighborhood structure during last $Itr$ iterations. $Suc_i$ stands for the successful time for solution improvement by using the $i$ th neighborhood structure during last $Itr$ iterations. Equation (18) normalizes the selection probabilities. It is expected that the neighborhood structure which has generated higher-quality solutions tends to increase its execution probability gradually. It is clear that the possibility for selecting each of the neighborhood structures is dynamic and self-adjustment.

5.2 Adaptive temperature control

The temperature and its declining pattern are adjusted to take the control of the SA behavior [23]. To avoid getting trapped in local minimum, worse moves might be accepted depending on the temperature which is gradually decreased under a procedure called cooling schedule as the algorithm proceeds [24]. As a lower temperature contributes to a lower transition probability, if a monotonic cooling schedule is adopted, the end of the search has less chance of escaping local minima while in the beginning, the search is more likely to move out of local minimum solutions [25]. Several cooling schedules have been proposed, such as linear cooling scheduling, exponential cooling scheduling, geometric schedules and quadratic cooling schedules. For further information, the reader may refer to Hajek [26], and Thompson and Dowsland [27]. In [25], the author proposed an adaptive temperature control function as below

$$\lambda_i = \lambda_{\min} + \rho \cdot In(1+r_i) \tag{19}$$

where $\lambda_{\min}$ is the minimum value of the temperature, $\rho$ is a weight coefficient, and $r_i$ is the number of consecutive bad moves at iteration $i$ (related to a temperature). The initial value of $r_i$ is zero, thus the initial temperature $\lambda_0 = \lambda_{\min}$.

The values of $r_i$ in (19) are updated as follows. At any iteration, if the current neighborhood transformation degrades the solution, the counter $r_i$ is increased by 1. If the new solution has an equal cost, $r_i$ remains unchanged. Otherwise if the new solution has been improved, $r_i$ is instantly equal to zero.

$$r_i = \begin{cases} r_{i-1}+1 & \text{if } \Delta f_i > 0 \\ r_{i-1} & \text{if } \Delta f_i = 0 \\ 0 & \text{if } \Delta f_i < 0 \end{cases} \tag{20}$$

The above temperature decreasing mechanism has been demonstrated to be very efficient in solving job shop scheduling problems compared to Van Laarhoven's Simulated Annealing algorithm [28]. In our tests, we find that the temperature controlled by function (19) rises too fast with the increasing of $r_i$, weakening the exploitation capability of Simulated Annealing algorithm. As a result, in this study, we improve the temperature control in (19) by introducing an additional coefficient $\delta$ to adjust the temperature rising rate. The new temperature control function is described as follows.

$$\lambda_i = \lambda_{\min} + \rho \cdot In(1 + \frac{r_i}{\delta}) \tag{21}$$

5.3 Combination of tabu-list

Integrated with the cooling schedule and the transition probability, SA has the capability to escape from local optima in the solution search process. However, the lack of memory may be regarded as the main deficiency of this method [29]. In comparison, the tabu search algorithm, introduced by Glover [30], memorizes a list of forbidden moves (tabu-list) to avoid short-term cycling and re-visiting, which may lead to extra computational overhead without improving the quality of the solution. Therefore, researchers consider integrating tabu-list into SA to enhance the performance of SA. In fact, the combination of tabu-list and SA has been reported to be effective in many studies [25, 29, 31]. The application process of the tabu-list is similar to [25].

A neighboring solution is generated by aforementioned insertion and removal neighborhood structure or immigration neighborhood structure. For each accepted neighboring perturbation, there may be tasks removed out of the former solution. As we know, at each iterate of SA, if a new generated neighboring solution is better (with higher profit) than the old one, then the new generated neighboring solution will be accepted. If the new generated neighboring solution has worse objective fitness than the old one, the new solution can be accepted with a certain probability which is determined by the current temperature and the fitness difference between the new and the old solution. Once the new solution has been accepted, the tasks removed from the old solution will be added to the tabu-list. The length of tabu-list is denoted by $L$. If the tabu-list is full, the task that was added in the tabu-list at the very beginning will be freed firstly. Every time when we conduct the neighborhood transformation, tasks in the tabu-list are not allowed to be inserted into the new solution until it was relieved.

5.4 Initial solution generation

A promising initial solution is also important to guarantee the efficiency of SA. The procedure of generation of initial solution generation procedure in our study is described as follows.

**Step 1:** Sort tasks in $U$ with the descendent order according to their weights.

**Step 2:** Select a task $t_k$ with the highest weight in $U$ sequentially and use the procedure of the insertion and removal neighborhood structure to generate partial solution $S'$ from the current partial solution $S$ step by step. If $S'$ is better than $S$, then $S'$ is accepted as the new current partial solution, i.e., $S \leftarrow S'$; otherwise, $S$ will not be updated. Remove $t_k$ out of $U$.

**Step 3:** Repeat Step 2 until all tasks in $U$ have been traversed.

From the discussions above, we provide the framework of ASA-DTC in **Algorithm 1**.

**Algorithm 1:** The Framework of ASA-DTC
1. Initialize the parameters $\lambda_{min}$, $\rho$, $\delta$, $\eta$, $Itr$, $pro_1$, $pro_2$, $L$.
2. Set $Sel_0 = 0$, $Suc_0 = 0$, $r_0 = 0$.
3. Generate initial solution $S$;
4. Let $S^*$ be the global elite solution, $S^* = S$;
5. **While** $g < MaxItr$ **do**
6.     $\lambda_g = \lambda_{min} + \rho \cdot In(1 + r_g/\delta)$
7.     $g = g + 1$
8.     Select the $i$ th neighborhood structure by roulette wheel method according to $pro_i$ to generate new solution $S'$;
9.     $Sel_i = Sel_i + 1$;
10. **If** $f(S) < f(S')$ **then**
11.     $S = S'$;
12.     $Suc_i = Suc_i + 1$;
13.     $r_g = 0$;
14.     If $tabuList$ is full, free the tasks earliest added in;
15.     Add the removed tasks from $S$ into $tabuList$;
16.     **If** $f(S) < f(S^*)$ **then**
17.         $S^* = S$;
18.     **End if**
19. **Else if** $\exp((f(S') - f(S))/\lambda_g) > \xi$, $\xi = uniform\ (0,1)$ **then**
20.     $S = S'$
21.     If $tabuList$ is full, remove the tasks earliest added in;
22.     Add the removed tasks from $S$ into $tabuList$;
23.     $r_g = r_{g-1} + 1$;
24. **End if**
25. **If** ($g\%Itr == 0$) **then**
26.     **For** $j = 1 \to 2$ **do**
27.         $pro'_j = \eta \cdot pro_j + (1-\eta) \cdot (Suc_j/Sel_j)$
28.     **End for**
29.     **For** $j = 1 \to 2$ **do**
30.         $pro_j = pro'_j / \sum_{j=1,2} pro'_j$
31.     **End for**
32.     $Suc_j = 0$;
33.     $Sel_j = 0$;
34. **End if**
35. **End while**
36. **Return** $S^*$.

## 6. Experimental studies

6.1 Simulation setup

To evaluate the performance of ASA-DTC, two groups of reference scenarios were designed for experimental studies. The difference between the first group and the second one is that targets contained in the second group are distributed among a much smaller area. The first group could represent the normal situation. The second group reflects some emergent situations such as earthquake, forest fire, flooding and military conflict. Meanwhile, the second group was used to test if the dynamic task clustering strategy contained in ASA-DTC would be particularly efficient when targets are densely distributed in a certain area. These two groups of reference scenarios are concisely described below.

In both these two groups, there are four on-orbit satellites of China, i.e. JB-3A, JB-3C, CBERS-1 and CBERS-2 (CBERS series satellites were cooperatively developed by China and Brazil). Each satellite circles the Earth about 100 minutes each time and runs about 14 orbits a day. Sensors on satellites can slew laterally among angle range $[-33^0, 33^0]$. Each group includes ten scenarios with 100 to 1000 point targets with a step of 100. Point targets in the first groups of scenarios randomly spread on the Earth surface with latitude among $[-30^0, 60^0]$ and longitude among $[0^0, 150^0]$. In comparison, point targets in the second group of scenarios are distributed on the Earth surface with latitude among $[30^0, 60^0]$ and longitude among $[90^0, 120^0]$. Each target is associated with a weight uniformly distributed among interval [2, 10]. The scheduling horizon is 24 hours. Time-windows and slewing angles associated with targets and satellites were calculated in prior using Analytical Graphics Inc.'s professional software named Satellite Tool Kit, also known as the STK (see www.stk.com). The orbit data of JB-3A, JB-3C, CBERS-1 and CBERS-2 can be downloaded from the satellite database of STK.

Other attributes of satellites and sensors and related parameters used in our experimental tests are listed in Table 1.

Table 1 Parameters used in the experimental setup

| $\sigma$ | $\Delta T$ | $W_j$ | $w_j$ | $E_j$ | $eo_j$ | $es_j$ | $v_j$ | $a_j$ | $c_j$ |
|---|---|---|---|---|---|---|---|---|---|
| 6 | 120 | 1000 | 1 | 1500 | 1 | 1 | 1 | 10 | 10 |

The algorithm was coded in C++ and run on a PC with Intel Pentium Dual E2108 (2.00 GHz) CPU and 1.0 GB RAM under Windows XP.

6.2 Computational results

In general, choosing proper parameters for the intelligent algorithm like Simulated Annealing is always time-consuming since parameters are always related. As a result of extensive simulations, the parameters of the ASA-DTC algorithm are specified in Tabel 2. Note that, in $L = N/50$, $N$ is the number of the targets in each experimental scenario.

Table 2 Parameters of the ASA-DTC algorithm

| $\lambda_{min}$ | $\rho$ | $\delta$ | $\eta$ | $Itr$ | $pro_1$ | $pro_2$ | $L$ |
|---|---|---|---|---|---|---|---|
| 0.5 | 1 | 10 | 0.8 | 10 | 0.5 | 0.5 | $N/50$ |

To assess the effectiveness of the ASA-DTC, we compared it with the Russian Doll Search (RDS) algorithm [1, 32], the Partition-based Approach (PBA) [13] and the Ant Colony Optimization with local search [10]. RDS is an algorithm producing exact solutions based on Branch-and-Bound techniques which can generate good results for small scale scheduling instance of Spot 5. PBA follows the divide and conquer principle and integrates dynamic programming with Tabu Search and can produce tight upper bound for scheduling problems of Spot 5. In fact, RDS, PBA and ACO-LS were integrated with the static task clustering strategy when applied to solving SOSPs. To make a difference between the static task clustering and the dynamic task clustering, we use PBA-STC, RDS-STC, and ACO-LS-STC to denote these three algorithms combined with the static task clustering strategy, respectively.

It should be noted that the scheduling problem considered in this paper is somewhat different from that of Spot 5. Some modifications were conducted to employ RDS and PBA. (1) Since one optical sensor was mounted on each satellite, stereo imaging requests were not considered and the so-called ternary constraints in were not taken into account. (2) Besides the memory storage constraint, the analogous knapsack constraint concerning energy capacity was considered, which was relaxed like the memory storage constraint when applied PBA.

The results for the first and the second group of scenarios are listed in Table 3 and 4, respectively. The obtained best results of each scenario are shown in boldface. In table 3 and 4,

$SN$ denotes the serial number of each scenario.

$TN$ denotes the number of time-windows between point targets and satellites.

$EN$ denotes the number of targets owning at least one time-window.

$R$ and $F$ denote the obtained profit and scheduled tasks by the RDS-STC algorithm.

$R^*$ and $F^*$ denote the upper bounds obtained by PBA-STC.

$\bar{R}$ respectively denote the mean profit obtained by ACO-LS-STC or ASA-DTC. Each algorithm ran for 50 times to solve the scheduling problem of each scenario.

$\bar{F}$ denotes the average number of the finished tasks.

$SD$ denotes the standard deviation of the ten profits obtained by running ACO-LS-STC or ASA-DTC for ten times to solve each case.

$\bar{T}$ denotes the mean running time.

$\overline{IMP}$ denotes the rate of profit improvement generated by ASA-DTC, compared to ACO-LS-STC.

In the last column, we show *t*-test results of a one-tailed *t*-test at a 0.05 level of significance to statistically assess results produced by ACO-LS-STC and ASA-DTC when solving each case. "+" means that ASA-DTC is significantly better (difference are statistically significant) than ACO-LS-STC.

Table 3 Simulation results of the first group of scenarios with targets distributed in a large area

| SN | EN | TN | PBA-STC | | RDS-STC | | ACO-LS-STC | | | | ASA-DTC | | | | $\overline{IMP}$ | t-test |
|---|---|---|---|---|---|---|---|---|---|---|---|---|---|---|---|---|
| | | | $R^*$ | $F^*$ | $R$ | $F$ | $\bar{R}$ | $\bar{F}$ | SD | $\bar{T}$ (s) | $\bar{R}$ | $\bar{F}$ | SD | $\bar{T}$ (s) | | |
| B1 | 92 | 227 | 565 | 94 | **556** | **92** | **556** | **92** | 0.0 | 105.51 | **556** | **92** | 0.0 | 135.62 | 0.0 | |
| B2 | 179 | 541 | 1097 | 169 | **1084** | **166** | 1082.1 | 165 | 0.32 | 193.48 | 1082.5 | 165 | 0.53 | 189.42 | 0.001 | + |
| B3 | 285 | 799 | 1679 | 265 | _ | _ | 1648.6 | 258 | 4.70 | 295.62 | **1656.2** | **258** | 5.61 | 265.35 | 0.005 | + |
| B4 | 381 | 1006 | 2199 | 358 | _ | _ | 2124.5 | 344 | 3.28 | 419.38 | **2142.7** | **348** | 7.78 | 454.61 | 0.009 | + |
| B5 | 441 | 1267 | 2567 | 391 | _ | _ | 2487.8 | 377 | 6.39 | 557.88 | **2517.8** | **385** | 8.51 | 662.57 | 0.012 | + |
| B6 | 512 | 1421 | 2924 | 443 | _ | _ | 2772.3 | 421 | 9.42 | 754.40 | **2816.3** | **429** | 8.67 | 886.78 | 0.015 | + |
| B7 | 603 | 1723 | 3404 | 516 | _ | _ | 3182.5 | 479 | 7.71 | 987.61 | **3228.3** | **490** | 10.61 | 1389.52 | 0.014 | + |
| B8 | 687 | 2064 | 3763 | 568 | _ | _ | 3597.4 | 544 | 14.58 | 1327.28 | **3662.8** | **559** | 11.63 | 1913.56 | 0.018 | + |
| B9 | 752 | 2238 | 4171 | 621 | _ | _ | 3956.2 | 586 | 14.79 | 1735.78 | **4042.4** | **601** | 15.17 | 2602.43 | 0.022 | + |
| B10 | 834 | 2516 | 4626 | 659 | _ | _ | 4338.4 | 612 | 16.40 | 2427.85 | **4441.5** | **639** | 17.01 | 3467.34 | 0.024 | + |

Table 4 Simulation results of the second group of scenarios with targets distributed in a small area

| SN | EN | TN | PBA-STC | | RDS-STC | | ACO-LS-STC | | | | ASA-DTC | | | | $\overline{IMP}$ | t-test |
|---|---|---|---|---|---|---|---|---|---|---|---|---|---|---|---|---|
| | | | $R^*$ | $F^*$ | $R$ | $F$ | $\bar{R}$ | $\bar{F}$ | SD | $\bar{T}$ (s) | $\bar{R}$ | $\bar{F}$ | SD | $\bar{T}$ (s) | | |
| C1 | 89 | 217 | 483 | 72 | 451 | 67 | 451 | 67 | 0.0 | 115.67 | **457** | **68** | 0.0 | 142.25 | 0.013 | + |
| C2 | 182 | 547 | 938 | 141 | 885 | 131 | 875.4 | 128 | 2.17 | 168.24 | **896.6** | **132** | 2.46 | 176.54 | 0.024 | + |
| C3 | 276 | 782 | 1455.6 | 216 | _ | _ | 1332.5 | 205 | 3.85 | 325.53 | **1389.8** | **211** | 4.21 | 285.35 | 0.043 | + |
| C4 | 385 | 1021 | *1621.5* | 249 | _ | _ | 1576.4 | 236 | 4.68 | 419.38 | **1655.2** | **247** | 4.36 | 494.61 | 0.050 | + |
| C5 | 450 | 1292 | *1795.2* | 268 | _ | _ | 1729.2 | 255 | 5.21 | 576.54 | **1809.2** | **269** | 5.87 | 705.62 | 0.046 | + |
| C6 | 504 | 1404 | *2018.4* | 291 | _ | _ | 1923.3 | 275 | 8.65 | 754.40 | **2043.4** | **291** | 9.52 | 872.54 | 0.062 | + |
| C7 | 611 | 1747 | 2329.6 | 349 | _ | _ | 2149.8 | 328 | 9.43 | 1114.57 | **2307.6** | **341** | 12.54 | 1469.48 | 0.073 | + |
| C8 | 675 | 2032 | 2721.3 | 391 | _ | _ | 2478.7 | 366 | 10.12 | 1327.28 | **2676.5** | **385** | 14.54 | 2023.71 | 0.080 | + |
| C9 | 755 | 2254 | *3003.8* | 445 | _ | _ | 2824.9 | 412 | 14.86 | 1907.34 | **3035.4** | **443** | 17.26 | 2744.65 | 0.076 | + |
| C10 | 865 | 2616 | *3324.3* | 487 | _ | _ | 3065.5 | 445 | 17.05 | 2827.85 | **3324.8** | **488** | 21.08 | 3654.45 | 0.085 | + |

From the obtained results listed in Table 3 and Table 4, we can easily arrive at some interesting observations.

(1) For each of these two groups of scenario, ASA-DTC always acquired better mean results than ACO-LS-STC did. Moreover, it is observed from the results of the *t*-test that, except B1 case, ASA-DTC is significantly better than ACO-LS-STC. This is because in ACO-LS-STC, the static task clustering and the task scheduling operation are separated into two phases, which influence the reasonability of the task clustering operation of the first phase without carefully considering the task scheduling situation in the second phase. In contrast, the dynamic task clustering strategy generates cluster-tasks dynamically during the task scheduling process, which is more flexible and fulfills the merit of task clustering to the highest extent.

(2) With the increase of the number of targets, the value of *IMP* increases accordingly, indicating that ASA-DTC shows more competitive performance in dealing with large scale SOSPs.

(3) The percentage of finished tasks in the first group of scenarios is higher than that in the second group. In addition, the values of *IMP* obtained in the second group of scenarios are significant higher than that of the first group. This can be explained that when immense numbers of targets densely spread in a relatively small area (like targets generated in the second group of scenarios), there will exist more conflicts among different observation tasks, which causes more tasks to fail in competing for the opportunity of being arranged into the schedule and therefore reduces the percentage of finished tasks. On the other hand, task clustering can alleviate the conflicts in a certain degree by enabling the formerly exclusive tasks to be completed simultaneously. In particular, the dynamic task clustering optimizes the task clustering process and gives the full scope to this potential, which thereby can remarkably improve the results. More interesting, ASA-DTC even obtained better results than PBA-STC did for scenario C4, C5, C6, C9 and C10.

(4) We can find that the running time of ASA-DTC on cases increases in an approximately linear trend as the number of targets getting larger. In addition, the relatively small standard deviations of the obtained results show the robustness of ASA-DTC in solving SOSPs. All these demonstrate that ASA-DTC is suitable for the large size real-world satellite observation scheduling problems.

6.3 Comparison with other scheduling algorithms

We also comprehensively compared ASA-DTC with other algorithms that combine the static task clustering strategy into ACO-LS (ACO-LS-STC) [10], typical Ant Colony Optimization [10], Tabu Search (TS-STC) [9], Genetic Algorithm (GA-STC) [7], Simulated Annealing (SA-STC) [5], a Highest Priority First Schedule algorithm (HPFS-STC). Concise descriptions of the comparative algorithms are given below.

ACO-LS: Ant colony optimization with local search strategy.

ACO: Traditional ant colony optimization does not integrate with the local search technique.

TS: The Tabu Search algorithm developed by Vasquez and Hao [9] was used to address the daily photograph scheduling of Spot 5. Some modification was implemented to cater to the difference between our scheduling problem and M. Vasquez's: (1) Removed the ternary constraints and related operations; (2) Added two knapsack constraints concerning energy capacity and maximum sensor slewing times.

GA: In literature [7], five criteria (i.e. priority, deadline, profit, area and emergency) were considered in the fitness function of GA. We considered only the priority criterion into the fitness function when applied the GA algorithm.

SA: We adopted the simulated annealing algorithm incorporating the mutation operator called temperature-dependent swap, which performed best in Globus's tests.

HPFS: This algorithm means highest priority first schedule. We rank observation tasks in descendent order with respect to their priorities and each time try to insert the task with the highest priority into the schedule

We are also interested in knowing about the performance of the proposed adaptive Simulated Annealing algorithm. For fairness, the adaptive Simulated Anneal algorithm with the static clustering strategy (ASA-STC) is also compared in the experiments. The comparison results are displayed in Fig 2.

It can be observed that (1) ASA-DTC is superior to all other peer scheduling algorithms. (2) ASA-STC is slightly better than TC-STC and significant better than SA-STC. As Tabu Search [9] can produce fairly good results for the scheduling problem of Spot 5 satellites, we say the adaptive mechanisms utilized to improve the conventional Simulated Annealing is indeed effective. (3) The naive and straightforward HPFS-STC algorithm exhibited much worse performance.

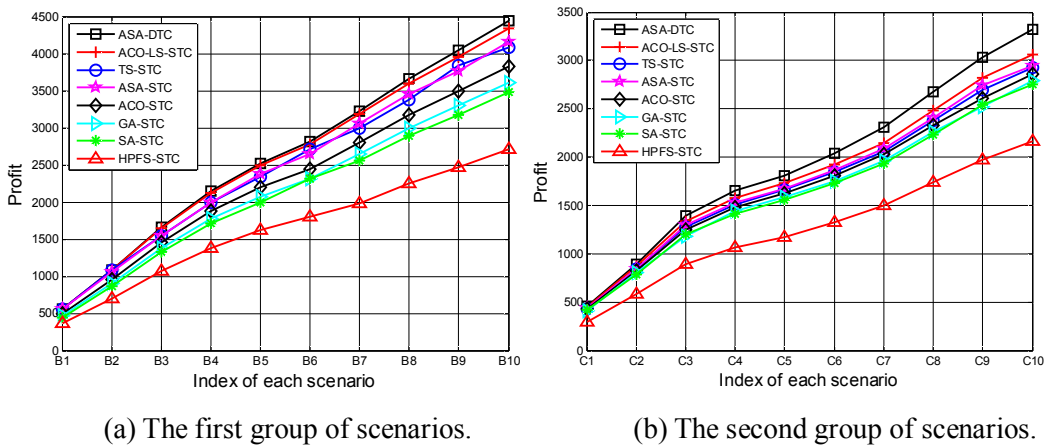

(a) The first group of scenarios.  (b) The second group of scenarios.

Fig.2. Comparison among different scheduling algorithms.

6.4. Effectiveness comparison between the dynamic task clustering strategy and the static task clustering strategy

From the data given in Table 3 and 4, and the related comparison result displayed in Fig. 2, we can see the effectiveness of the task clustering strategy, especially the dynamic task clustering strategy. To give an intuitive and straightforward illustration of the impact of different clustering strategies, in Fig.3, we compared algorithms of adaptive Simulated Annealing respectively combined with static task clustering strategy (ASA-STC), the dynamic clustering strategy (ASA-DTC) or without combining with any task clustering strategy (ASA-NONTC).

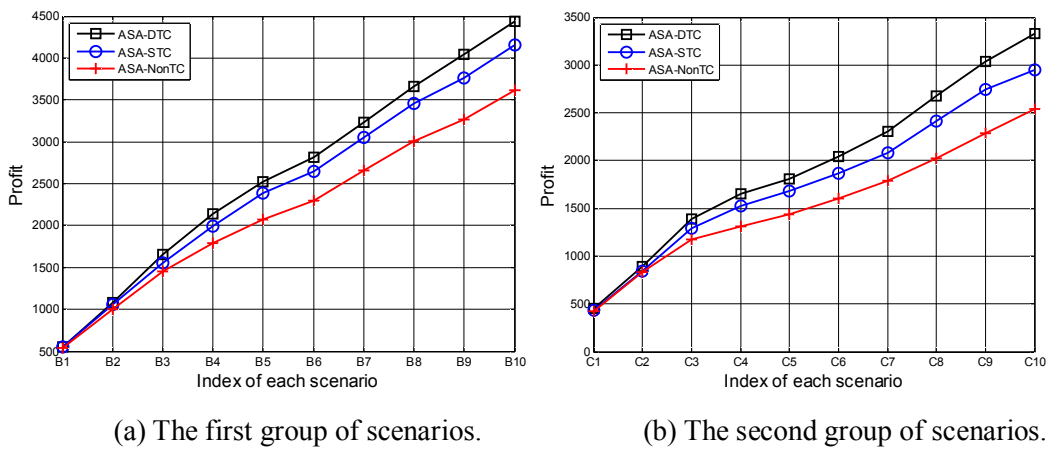

(a) The first group of scenarios.  (b) The second group of scenarios.

Fig.3. Comparison among dynamic task clustering, static task clustering and non-task clustering.

Some observations can be drawn from Fig. 3, (1) ASA-DTC and ASA-STC produced much better results than ASA-NONTC did, especially when the scheduling problem contained relatively large number of targets. This shows successful application of these two task clustering strategies. (2) ASA-DTC outperforms ASA-STC. Besides, we can find that with the number of targets getting larger, the superiority of ASA-DTC appears more significant. This indicates that the dynamic task clustering strategy is more effective than the static task clustering strategy. (3) Compared to solving the first group of scenarios, ASA-DTC exhibits even better performance than ASA-STC in solving the second group of scenarios. This phenomenon further proves that the dynamic task clustering strategy is particularly effective when solving SOSPs with a great quantity of targets being densely distributed. The reason is that exclusive conflicts among observation tasks included in such problems are heavy, while the dynamic task clustering strategy is more powerful and flexible to reduce these conflicts.

6.5 Impact of different task clustering strategies on cost-benefits

One may be concerned with on how different task clustering strategies impact on cost-benefits. In this study, the cost is related to the consumption of on-board energy and memory storage; and the benefit is related to the obtained profits. Intuitively, such impact is case-dependent. There are two situations. The first one is that time-window length of obtained cluster-tasks is longer than the sum of time-window length of all component tasks. In this case, task clustering may cause the satellite to consume on average more resources to obtain the same profit. In contrast, the second situation is that time-window length of obtained cluster-tasks is shorter than the sum of time-window length of all component tasks. In this case, task clustering could save resources. Intuitively, whether task clustering will save resources or not is depended on the percentage of cluster-tasks produced at the second situation.

To find some underlying relations between different task clustering strategies and the cost-benefit factor, we conduct numerical analyses of the experimental tests. Fig. 4 shows the ratio between the obtained profit and the consumed memory storage (profit-memory ratio) when utilized ASA-STC, ASA-DTC and ASA-NONTC to address each scenario. We can find from Fig.4 that (1) profit-memory ratios of ASA-NONT are usually higher than that of ASA-STC and ASA-DTC, stressing that to achieve the same profit, the task clustering strategies will on average utilize more memory storage. (2) With the number of targets increasing, the profit-memory ratios of all these three algorithms will become higher. This can be explained that when there are more observation tasks (i.e. targets) requesting for finishing, each algorithm inclines to choose tasks with higher weights to execute, which causes the rise of the ratio. (3) Profit-memory ratios of ASA-DTC are always larger than that of ASA-STC. In particular, when the number of targets getting larger or targets spread densely (related to the second group of scenarios) in an area, the ratio gap between ASA-DTC and ASA-STC becomes larger as well. This is resulted from the addition of the resource consumption constraint into the dynamic task clustering process of ASA-DTC.

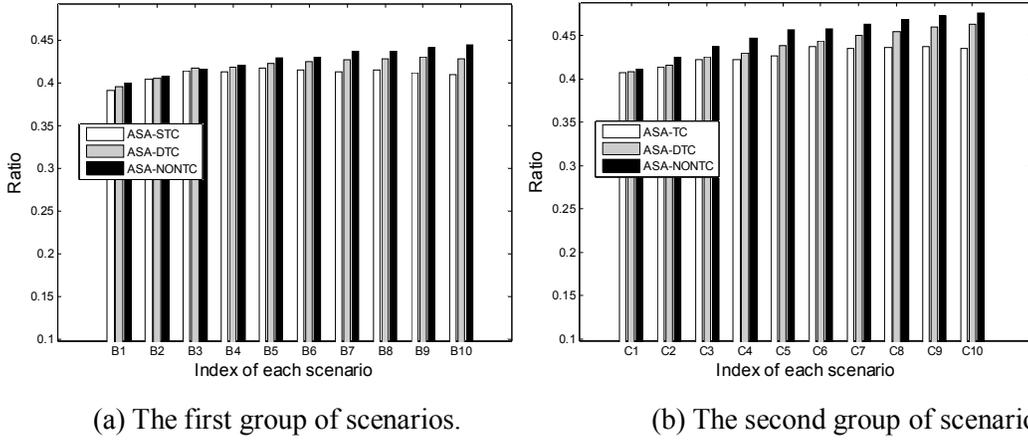

(a) The first group of scenarios.   (b) The second group of scenarios.

Fig.4. Comparison of the ratio between the obtained profit and the consumed memory storage.

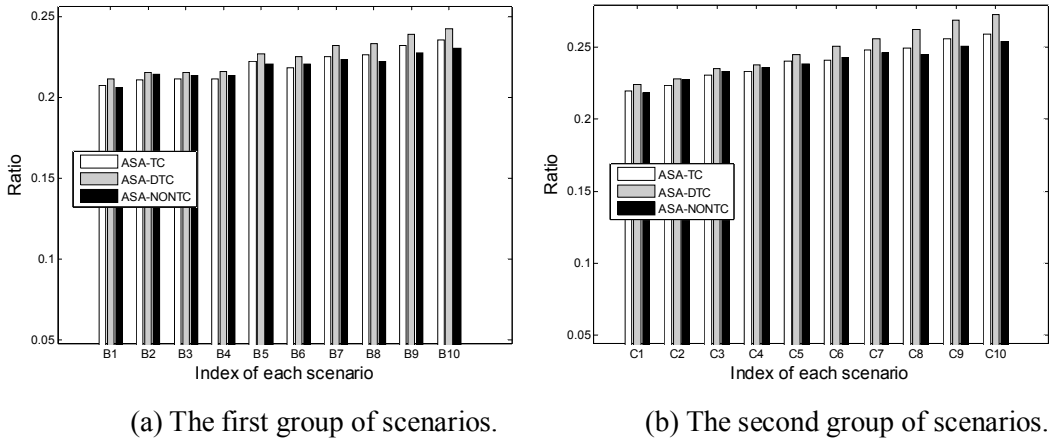

(a) The first group of scenarios.   (b) The second group of scenarios.

Fig.5. Comparison of the ratio between the obtained profit and the consumed energy.

Similarly, Fig. 5 compares the ratio between the obtained profit and the consumed energy (profit-energy ratio) when employing ASA-STC, ASA-DTC and ASA-NONTC, respectively. It can be observed from Fig. 5 that (1) except in scenario B2, B3 B4, C2 C3 and C4, profit-energy ratios of ASA-DTC and ASA-STC are higher than that of ASA-NONTC, indicating that task clustering strategies have the potential to save energy. This is because on the one hand task clustering strategies can save resources, on the other hand, task clustering always means less sensor opening times and smaller sensor slewing angles, which will effectively save energy. (2) Similar to profit-memory ratios and for the same reason, profit-energy ratios of all these three algorithms also rise with the scale of targets getting larger. (3) Profit-energy ratios of ASA-DTC are always higher than that of ASA-STC and ASA-NONTC, especially in the case of a higher number of targets or these targets are closely distributed (related to the second group of scenarios). This further underlines the effectiveness of the dynamic task clustering strategy and the resource consumption constraint.

## 7. Conclusions

The consideration of tasks clustering can improve the scheduling efficiency of SOSPs. To overcome the deficiency of the static task clustering, in this study, we propose a dynamic task clustering strategy, which is integrated into an adaptive Simulated Annealing scheduling method, such that the ASA-DTC scheduling algorithm for SOSPs is developed. In ASA-DTC, task clustering operations are incorporated into the neighborhood structures and performed dynamically as ASA-DTC proceeds. In addition, several adaptive mechanisms are adopted in the ASA-DTC, including an adaptive temperature control, a tabu-list based revisiting avoidance technique, and the intelligent combination of neighborhood structures, to improve the typical Simulated Annealing algorithm.

Extensive experiments and analysis demonstrate that ASA-DTC produces better results for each simulated scenario of SOSP within reasonable computation cost. ASA-DTC shows superior performance in the experimental studies when compared with some other state-of-the-art satellite observation scheduling algorithms, such as ACO-LS, ACO, TS, GA, SA and HPFS. Task clustering strategies are again shown to be very effective to improve the scheduling efficiency of SOSPs. In particular, when compared to the static task clustering strategy, the dynamic task clustering strategy is significantly better in facilitating the scheduling algorithm to produce higher-quality schedules and save resources, i.e., energy and memory storage. ASA-DTC is expected to be especially useful when SOSPs contain a large number of targets or the targets are closely distributed. In addition, the new adaptive Simulated Annealing algorithm is expected to be useful in other scheduling problems, such as job shop scheduling and flowing shop scheduling.

The future research can be carried in three directions. (1) Application of the proposed algorithm to scheduling of agile satellites. (2) Considering satellite observation and data transmission simultaneously when scheduling satellite activities. (3) Designing dynamic and robust scheduling approaches to adapt to unexpected situations, such as emergency tasks, requirement change of tasks, malfunction of satellites and environment disturbances (e.g., cloud).

## Acknowledgements

This work was supported by the National Nature Science Foundation of China under Grant No. 71271213. The Author Guohua Wu is supported by the China Scholarship Council under Grant No.201206110082.